\documentclass{article}

\usepackage{url}
\usepackage{microtype}
\usepackage{parskip}
\usepackage[super]{natbib}
\usepackage[a4paper, left=2.5cm, right=2.5cm, top=2.5cm, bottom=2.5cm]{geometry}
\usepackage{longtable,booktabs}
\usepackage{caption}
\usepackage{blindtext}
\usepackage{graphicx}
\usepackage{authblk}
\usepackage{amsmath}
\usepackage{lineno}
\usepackage[toc,page]{appendix}
\usepackage[utf8]{inputenc}

\usepackage[ruled,vlined]{algorithm2e}

\usepackage{multicol}
\title{Integrating Deep Reinforcement Learning Networks with Health System Simulations.\\
\vspace{0.5cm}
\large{A practical OpenAI Gym-like coding framework using PyTorch and SimPy.}}

\author[1]{Michael Allen}
\author[2]{Thomas Monks}

\affil[1]{\footnotesize University of Exeter Medical School \& NIHR South West Peninsula Applied Research Collaboration (ARC).}
\affil[2]{\footnotesize University of Exeter Institute of Data Science and Artificial Intelligence}

\begin{document}

\maketitle

\section*{Abstract} 

\emph{Background and motivation}: Combining Deep Reinforcement Learning (Deep RL) and Health Systems Simulations has significant potential, for both research into improving Deep RL performance and safety, and in operational practice. While individual toolkits exist for Deep RL and Health Systems Simulations, no framework to integrate the two has been established. 

\emph{Aim}: Provide a framework for integrating Deep RL Networks with Health System Simulations, and to ensure this framework is compatible with Deep RL agents that have been developed and tested using OpenAI Gym.

\emph{Methods}: We developed our framework based on the OpenAI Gym framework, and demonstrate its use on a simple hospital bed capacity model. We built the Deep RL agents using PyTorch, and the Hospital Simulation using SimPy.

\emph{Results}: We demonstrate example models using a Double Deep Q Network or a Duelling Double Deep Q Network as the Deep RL agent.

\emph{Conclusion}: SimPy may be used to create Health System Simulations that are compatible with agents developed and tested on OpenAI Gym environments.

\emph{GitHub repository of code}: \url{https://github.com/MichaelAllen1966/learninghospital}

\begin{multicols}{2}

\section{Introduction}

Deep Reinforcement Learning and Health System Simulations are two complementary and parallel methods that have the potential to improve the delivery of health systems.

Deep Reinforcement Learning (Deep RL) is a rapidly developing area of research, finding application in areas as diverse as game playing, robotics, natural language processing, computer vision, and systems control \cite{li_deep_2018}. Deep RL involves an \emph{agent} that interacts with an \emph{environment} with the aim of developing a \emph{policy} that maximises long term \emph{return} of \emph{rewards}. Deep RL has a framework that allows for generic problem solving that is not dependent on pre-existing domain knowledge, making these techniques applicable to a wide range of problems.

Health Systems Simulation seeks to mimic the behaviour of real systems. These may be used to optimise services such as emergency departments \cite{monks_using_2017}, hospital ward operation and capacity \cite{penn_towards_2020} and community hospital capacity  \cite{monks_modelling_2016}. These examples of health service simulations are used for off-line planning and optimization of service configuration. 

Health Systems simulations are usually used for planning of service delivery changes. There is potential for these type of simulations to be used to test, develop and train Deep RL agents. The motivation for this integration includes:

\begin{itemize}

    \item To perform research on the relative performance of different Deep RL methods (e.g. comparison of techniques such as \emph{Deep Q Learning} and \emph{Actor-Critic} methods).
    
    \item To perform research on the effect of differing reward structures on the performance of Deep RL agents, and enabling the development of reward structures that carefully balance average performance with safety (avoiding rare but catastrophic events). 
    
    \item Ultimately, to be able to pre-train Deep RL agents which would then be transferred to, and used in, real world settings.

\end{itemize}

In order to test, train, and develop Deep RL agents, we need a standardised structure that we can use across different types of health systems. One such standardised structure, used across many differing domains, already exists, and that is OpenAI Gym (\url{gym.openai.com}) \cite{brockman_openai_2016}. Gym provides a common interface to a range of problems, from control systems through to video games. The common interface allows the easy transfer of agents from one problem-solving environment to another. Gym is structured on an episodic framework to learning. The agent is exposed to multiple iterations, where the environment is \emph{reset} to a fixed or random state, and the agent then interacts with the environment through a series of \emph{steps} until some \emph{terminal} state is reached indicating the end of the episode. With each step that agent passes an \emph{action} to the environment. The environment returns an updated set of \emph{observations} about the environment \emph{state}, a reward, whether the terminal state has been reached, and any extra information available. Agents are designed to maximise the return of long term rewards.

In this paper we present a framework for coding Health Systems simulations, using the commonly used Python discrete event simulation package, SimPy \cite{team_simpy_simpy_2020} in a framework that allows interaction with a Deep RL agent. The framework uses RL agent method calls with high compatibility with OpenAI Gym, allowing easy transfer of agents developed with/for OpenAI Gym environments. As a demonstration, we use a simple hospital bed simulation model in SimPy, and show interaction with two Deep RL agents (written with PyTorch): a Double Deep Q Learning Network and a Duelling Deep Q Network. Our intention is not to provide a robust hospital bed simulation model, nor to provide an optimised Deep RL agent for such use, but to demonstrate a framework for combining Gym-compatible Deep RL agents with Health Systems simulations in SimPy.

\section{GitHub repository}

The GitHub repository containing this code is:

\url{https://github.com/MichaelAllen1966/learninghospital}

The examples cited in this paper are from release version 1.0.0 (DOI 10.5281/zenodo.3936515):

\url{https://github.com/MichaelAllen1966/learninghospital/releases/tag/v0.0.1}.

\section{Method}

\subsection{Generic simulation properties}

All simulations will share some common structure, methods, and attributes.

\subsubsection{Generic structure}

Algorithm \ref{algo:overview} shows a high level structure of the code. This will be common to all interactions of Deep RL agents and SimPy simulations with only RL-specific alterations (such as the use of \emph{target networks} and \emph{memory}).

\begin{algorithm}[H]
\caption{High level view of model using A Double Deep Q Network (using policy net, target net, and memory)}
\SetAlgoLined
Set up policy net\;
Set up target net\;
Set up memory\;
\While{Training episodes not complete}{
    Reset sim\;
    \While{not in terminal state}{
        Get action from policy net\;
        Pass action to sim\;
        Take a time-step in sim\;
        Receive (next state, reward, terminal, info) from sim\;
        Add (state, next state, reward, terminal) to memory\;
        Render environment (optional)\;
        Update policy net;\
        }
    Update target net\;
    }
Assess performance of policy net\;
\label{algo:overview}    
\end{algorithm}

\subsubsection{Generic simulation methods}

The simulation is set up with three methods that interface the Deep RL agent and the simulation:

\begin{itemize}
    \item \emph{reset}: resets the sim to a starting state and returns the first set of state observations.
    \item \emph{step}: takes a step in the simulation. Passes an action to the simulation. Runs the simulation until the end of the next time step, and returns a tuple of next state, reward, terminal, info. The \emph{step} method uses the SimPy method \emph{env.run(until=target-time)}, with target-time being incremented in the desired time steps. When the simulation time reaches the desired maximum simulation duration, the simulation returns \emph{terminal=True}.
    \item \emph{render}: displays the current state of the simulation.
    
\end{itemize}

Other internal methods in the simulation (not accessed by the Deep RL agent) that will be common to all simulations are:

\begin{itemize}
    \item \emph{calculate\_reward}: calculates the reward to pass back to the Deep RL agents.
    \item \emph{get\_observations}: creates a \emph{list} of observations from the state.
    \item \emph{islegal}: checks whether an action from the Deep RL agent is legal. If the action is not legal, this method will raise an exception.
\end{itemize}

\subsubsection{Generic simulation attributes}

All simulations will contain the following attributes.

\begin{itemize}
    \item \emph{actions}: A list of possible actions.
    \item \emph{action\_size}: The number of possible actions.
    \item \emph{observation\_size}: The number of features in the observation.
    \item \emph{state}: An object containing the state of the simulation. This may be a simple object, such as a list or dictionary, or may be a custom Python object.
\end{itemize}

\subsection{Hospital bed simulation}

\subsubsection{Hospital bed simulation overview}

The hospital bed simulation is a very simplified model of a real hospital. Patients arrive at a hospital, stay for a given length-of-stay, and leave. The inter-arrival time of patients is sampled from an exponential distribution, the mean of which depends on the day of week (with average arrival numbers being higher on weekdays than weekends). The length-of-stay is also sampled from an exponential distribution, the mean of which does not depend on day of week. The hospital has a certain number of beds at any time. The Deep RL agent can request a change to the number of staffed beds, but this change is only enacted after 2 days. The simulation runs for 365 days by default, and the hospital is loaded initially with the expected average number of patients.

\subsubsection{Hospital bed simulation \emph{state}}

The state in the simulation is held by a dictionary. This dictionary contains:

\begin{itemize}
    \item \emph{weekday}: The current day of week (0-6).
    \item \emph{beds}: The total number of staffed beds in the hospital (free or occupied).
    \item \emph{patients}: The total number of patients in the hospital.
    \item \emph{spare\_beds}: The number of unoccupied beds. If the number of patients exceeds the number of staffed beds then this number becomes negative and indicated the number of patients without a bed.
    \item \emph{pending\_bed\_change}: The changes in staffed bed numbers requested by the Deep RL agent, but which has not yet been actualised.
\end{itemize}

\subsubsection{Hospital bed simulation reward}

The simulation has a target number of free staffed beds. By default this is set at 5\% the number of patients in the hospital at any given time. The \emph{reward} is always zero or negative and is the negative difference between the number of spare beds and the target number of spare beds (equation \ref{eq_1}).

\begin{small}
\begin{equation}
    reward = -abs(spare\ beds-target\ spare\ beds)
\label{eq_1}
\end{equation}
\end{small}

\subsubsection{Hospital bed simulation methods}

Methods that are specific to the hospital bed simulation are:

\begin{itemize}
    \item \emph{adjust\_bed\_numbers}: Adjusts the staffed bed numbers after a delay (SimPy \emph{timeout}). Prior to the delay, the \emph{adjust\_pending\_bed\_change} method is called to track the requested changes in staffed bed numbers. The delay is the simulation time between the Deep RL agent requesting a change to the number of staffed beds, and the change being made. The delay is stored in the simulation attribute \emph{delay\_to\_change\_beds}, and may be set when initializing the simulation. When the number of staffed beds changes, the state dictionary items \emph{beds} and \emph{pending\_bed\_change} are adjusted accordingly.
    
    \item \emph{adjust\_pending\_bed\_change}: Adjusts the state dictionary item \emph{pending\_bed\_change} when the Deep RL agent requests a change to the number of staffed beds. 
    
    \item \emph{load\_patients}: Loads new patients at the start of the simulation, such that the initial number of patients in the hospital equals the calculated long term average ($arrivals\ per\ day * average\ length\ of\ stay)$. This method calls the \emph{patient\ spell method}. This method increments the number of patients and staffed beds by 1 for each patient loaded into the simulation.
    
    \item \emph{new\_admission}: A continuous loop of new patients. This method/process is initiated on simulation reset. A new patient arrival is initiated by calling the \emph{patient\_spell} method. The number of patients in the hospital is incremented by 1. There is then a delay (SimPy \emph{timeout}) before the next iteration of the loop. The delay is the inter-arrival time of patients. This is sampled from an exponential distribution, the mean of which depends on both the average arrival rate (set using \emph{arrivals\ per\ day} attribute, which may be set when initializing the simulation. Mean arrivals per day are increased by 20\% on weekdays (days 0-4), and reduced by 50\% on weekends (days 5 \& 6).
    
    \item \emph{patient\_spell}: The patient spell in the hospital. Length of stay is sampled from an exponential distribution based on a mean length of stay. If the patient is part of the initial load of the hospital, the length of stay is multiplied by a random number between 0-1 to account for the fraction of the length of stay already completed. After the spell in hospital is complete the number of patients is reduced by 1, and the number of spare beds recalculated.
\end{itemize}

\subsubsection{Hospital bed simulation reset method}

The actions in the simulation \emph{reset} method (required in all simulations for interaction with the Deep RL agents) are:

\begin{enumerate}
    \item Create new hospital simulation environment.
    \item Initialise simulation processes (\emph{new\_admission} method).
    \item Set starting state values for state dictionary.
    \item Call \emph{load\_patients} method.
    \item Get and return first set of state observations.
\end{enumerate}

\subsubsection{Hospital bed simulation step method}

The actions in the simulation \emph{step} method (required in all simulations for interaction with the Deep RL agents) are:

\begin{enumerate}
    \item Check requested action is legal.
    \item Adjust pending bed change.
    \item Call bed change process.
    \item Make a step in the simulation.\\ Use: \emph{env.run(until=self.next\_time\_stop)}.
    \item Get new observations.
    \item Check whether terminal state reached (based on simulation time).
    \item Get reward.
    \item Create an empty information dictionary (this dictionary is required to be compatible with OpenAI Gym step method).
    \item Render environment if requested.
    \item Return tuple of next state, reward, terminal, info.
\end{enumerate}

\subsubsection{Hospital bed simulation attributes}

Attributes that are specific to the hospital bed simulation are:

\begin{itemize}
    \item \emph{arrivals\_per\_day}: Average arrivals per day.
    \item \emph{delay\_to\_change\_beds}: Time between requesting change in beds, and change in beds happening (days).
    \item \emph{los}: Average patient length of stay (days).
    \item \emph{sim\_duration}: Length of simulation run (days).
    \item \emph{target\_reserve}: target free staffed beds as a proportion of the number of patients present.
    \item \emph{time\_step}: Time between action steps (day).
\end{itemize}

\subsection{Deep Reinforcement Learning Agents}

A range of standard Deep RL agents were implemented for this model (and are provided in separate notebooks in the associated GitHub repository):

\begin{enumerate}
    \item \emph{Double Deep Q Network (D2QN)}: Standard Deep Q Network, with policy and target networks \cite{van_hasselt_deep_2015}.
    \item \emph{Duelling Double Deep Q Network (D3QN)} \cite{wang_dueling_2016}: Policy and target networks calculate Q from sum of *value* of state and *advantage* of each action (*advantage* represents the added value of an action compared to the mean value of all actions).
    \item \emph{Noisy D3QN} \cite{fortunato_noisy_2019}: Networks have target layers that add Gaussian noise to aid exploration.
    \item \emph{Prioritised replay D3QN} \cite{schaul_prioritized_2016}: When training the policy network, steps are sampled from the memory using a method that prioritises steps where the network had the greatest error in predicting Q.
    \item \emph{Bootstrapped D3QN} \cite{osband_deep_2016}: Multiple networks are trained from different bootstrap samples from the memory.
    \item Combinations of the above
\end{enumerate}

\section{Results}

The GitHub repository contains examples of the various Deep RL agents implemented.

\begin{figure}
\centering
\includegraphics[width=0.9\textwidth]{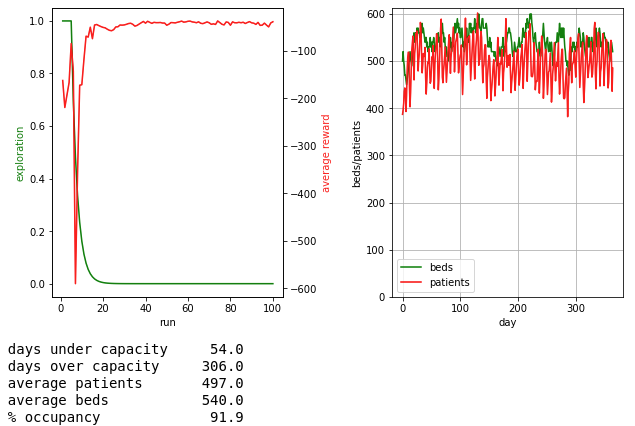}
\caption{Example model. A Bagging Duelling Double Deep Q Network agent trained to manage bed capacity of the hospital bed simulation. \emph{Left}: Exploration (epsilon, the probability of taking an action purely at random, green) and average reward across the training episode (red). \emph{Right}: The number of patients (red) and staffed beds (green) in the last training run}
\label{fig:results_dqn}
\end{figure}

The output of the Bagging D3QN is shown in figure \ref{fig:results_dqn}. It is not the intention of this paper to present a fully optimised Deep RL agent, but it can be seen that the example network improves in performance over time (repeated model runs) and manages the modelled bed stock appropriately.

\section{Discussion}

Combining Deep RL and Health Systems Simulations has significant potential, for both research into improving Deep RL performance and safety, and in operational practice. Our aim in this paper is not to present an optimised Deep RL model, or a detailed hospital simulation, but to provide a framework that is compatible with OpenAI Gym environments, enabling easy transfer of the many methods that have been developed and tested in such environments. 

The potential for combining Deep Learning and Health Systems Simulations goes beyond the framework provided here. For example, we have demonstrated that a machine learning model can be used to simulate patient-level clinical decision making \cite{allen_can_2019} as part of broader clinical pathway simulation study.

The combination of Deep Learning and Health Systems Simulation is an area of research that will hopefully bear much fruit in the coming years.

\end{multicols}

\begin{multicols}{2}

\section{References}
\bibliographystyle{ieeetr}
\bibliography{refs}
\end{multicols}

\section*{Funding}

This study was funded by the National Institute for Health Research (NIHR) Applied Research Collaboration (ARC) South West Peninsula. The views and opinions expressed in this paper are those of the authors, and not necessarily those of the NHS, the National Institute for Health Research, or the Department of Health.

\end{document}